\documentclass{article}

\usepackage[preprint]{neurips_2026}

\usepackage[utf8]{inputenc}
\usepackage[T1]{fontenc}
\usepackage{hyperref}
\usepackage{url}
\usepackage{booktabs}
\usepackage{amsfonts}
\usepackage{amssymb}
\usepackage{amsmath}
\usepackage{nicefrac}
\usepackage{microtype}
\usepackage{xcolor}
\usepackage{graphicx}

\newcommand{\teArrow}{\ensuremath{\rightarrow}}
\newcommand{\teImplies}{\ensuremath{\Rightarrow}}
\newcommand{\teTherefore}{\ensuremath{\therefore}}
\newcommand{\teBecause}{\ensuremath{\because}}
\newcommand{\teUp}{\ensuremath{\uparrow}}
\newcommand{\teDown}{\ensuremath{\downarrow}}
\newcommand{\teAnd}{\ensuremath{\wedge}}
\newcommand{\teOr}{\ensuremath{\vee}}
\newcommand{\teNot}{\ensuremath{\neg}}
\newcommand{\teApprox}{\ensuremath{\approx}}
\newcommand{\teNeq}{\ensuremath{\neq}}

\title{Telegraph English: Semantic Prompt Compression\\via Structured Symbolic Rewriting}

\author{%
  Mikhail L. Arbuzov \\
  Independent Researcher \\
  \texttt{Mike.arbuzov54@gmail.com} \\
  \And
  Sisong Bei \\
  Independent Researcher \\
  \texttt{qurining@gmail.com} \\
  \And
  Ziwei Dong \\
  Independent Researcher \\
  \texttt{ziwei.dong@alumni.emory.edu} \\
  \And
  Dmitri Kalaev \\
  Independent Researcher \\
  \texttt{kalaevdr@gmail.com} \\
  \And
  Alexey Shvets \\
  Palo Alto Networks \\
  \texttt{ashvets@paloaltonetworks.com} \\
}

\begin{document}

\maketitle

\begin{abstract}
We introduce Telegraph English (TE), a prompt-compression protocol that rewrites natural language into a symbol-rich, formally-structured dialect. Where token-deletion methods such as LLMLingua-2 train a classifier to delete low-importance tokens at a fixed ratio, TE performs a full semantic rewrite: it decomposes the input into atomic fact lines, substitutes verbose phrases with $\sim$40 logical and relational symbols, and lets the compression ratio adapt to each document's information density. A consequence of the line-structure rule is that compression and semantic chunking become the same operation---each output line is an independently addressable fact, so the compressed representation is simultaneously a semantic index. We evaluate TE on 4{,}081 question-answer pairs from LongBench-v2 across five OpenAI models and two difficulty levels. At roughly 50\% token reduction, TE preserves 99.1\% accuracy on key facts with GPT-4.1 and outperforms LLMLingua-2 at matched compression ratios on every model and task tested. The gap widens on smaller models---up to 11 percentage points on fine-detail tasks---suggesting that explicit relational structure compensates for limited model capacity. We release the grammar specification, compression prompt, benchmark data, and reference implementation.
\end{abstract}

\section{Introduction}
\label{sec:intro}

Large language models are increasingly embedded in retrieval-augmented generation (RAG), multi-agent orchestration, and long-context reasoning pipelines. Input cost scales linearly with token count, so prompt compression---feeding fewer tokens to the model while preserving the information it needs---has become a practical lever for controlling latency and cost.

Two families of approach exist. \emph{Extractive} methods select a subset of tokens or sentences from the input \citep{jiang2023llmlingua,pan2024llmlingua2}; \emph{abstractive} methods paraphrase or summarise it \citep{chevalier2023autocompressors}. LLMLingua-2 \citep{pan2024llmlingua2}, currently the strongest published baseline, trains a GPT-4-distilled XLM-RoBERTa-large classifier to delete tokens below an importance threshold at a user-specified ratio.

Token deletion works, but it has structural limits that become visible once one looks past the compression ratio. The ratio is fixed regardless of input density. Deleting tokens can sever co-reference chains and destroy logical connectives, leaving the downstream model to hallucinate the relationships between surviving fragments. Token-deletion methods are input-only preprocessors---they compress the initial prompt, but generated output passes uncompressed to the next pipeline stage, so multi-step agent systems cannot compound the savings. Most consequentially, token deletion produces no structure: the output is a degraded copy of the input, unable to be indexed, selectively pruned, or dynamically updated.

We propose Telegraph English (TE), a different kind of compression. Rather than selecting which tokens to keep, TE rewrites the passage into a compact, formally-structured dialect. The original sentence
\begin{quote}\small
\emph{``According to research by Johnson and colleagues (2023), the application of machine learning techniques to medical diagnostics resulted in a 27.5\% increase in early detection rates while simultaneously reducing false positives by approximately 12\% compared to traditional methods.''}
\end{quote}
becomes, under TE:
\begin{quote}\footnotesize\ttfamily
ML\teArrow{}MEDICAL-DIAGNOSTICS: EARLY-DETECTION+27.5\% \teAnd{} FALSE-POSITIVE-12\% [JOHNSON:2023]
\end{quote}
Sixty-eight tokens become fourteen. The causal relationship, both quantitative claims, and the citation are each on record as separate, addressable units---and the phrase ``application of\dots resulted in'' has collapsed into a single symbol.

What makes TE architecturally distinctive is a property that emerges from the grammar's line-structure rule: compression and semantic chunking are the same operation. Every TE output line contains exactly one atomic fact---one claim, one relationship, one datum. This is not a post-processing step but a consequence of how the grammar defines a legal output. The result is a representation that is simultaneously compressed, retrieval-ready, and amenable to dynamic management: atomic lines are individually embeddable; tagged sections support hierarchical context budgeting; and facts can be updated, merged, or pruned without re-running the compressor.

\paragraph{Contributions.}
(1)~A formal grammar specification for structured prompt compression (\S\ref{sec:grammar}). (2)~A unified compression-and-chunking framework where semantic compression, retrieval-ready indexing, and dynamic context management emerge from a single rewriting pass (\S\ref{sec:grammar}, Appendix~\ref{app:semantic-chunking}). (3)~A large-scale empirical comparison against LLMLingua-2 on 4{,}081 key-fact and 801 fine-detail QA pairs across five models (\S\ref{sec:results}). (4)~Evidence that the advantage of semantic rewriting over token deletion grows on smaller models and on detail-intensive tasks (\S\ref{sec:analysis}). (5)~A reference implementation with CLI tools for compression, benchmarking, and error analysis.

\section{Related Work}
\label{sec:related}

\paragraph{Prompt compression.}
LLMLingua \citep{jiang2023llmlingua} introduced budget-constrained prompt compression using perplexity-based token selection. LLMLingua-2 \citep{pan2024llmlingua2} improved on this with a data-distillation approach: GPT-4 labels token importance on the MeetingBank corpus, and an XLM-RoBERTa-large classifier learns to predict which tokens to delete. The compressor is domain-agnostic in principle, though Pan et al.\ note effectiveness decreases on domains with different token-importance distributions from the training data. The architectural constraint is that the output remains a degraded subset of the input tokens---no new structure is introduced.

\paragraph{Abstractive compression.}
AutoCompressors \citep{chevalier2023autocompressors} train summary tokens that substitute for long contexts; RECOMP \citep{xu2023recomp} generates abstractive summaries tailored to retrieval queries. Both are effective but lossy by design---they discard information that cannot be recovered, and neither produces a structured output that supports selective manipulation.

\paragraph{Structured representations and agent context.}
Chain-of-thought prompting \citep{wei2022cot} and structured prompting \citep{hao2023structured} demonstrate that imposing structure on LLM inputs improves reasoning. TE extends this insight to compression: explicit logical and relational operators help downstream models reconstruct the intended meaning more reliably than degraded natural language. For long-running agents, MemGPT \citep{packer2023memgpt} addresses context-window growth via virtual memory hierarchies, and Reflexion \citep{shinn2023reflexion} maintains explicit self-reflection buffers; both operate on natural-language representations. TE offers a complementary strategy of structured, fact-level representations that can be selectively updated and pruned without further LLM calls.

\paragraph{Semantic chunking for RAG.}
Standard RAG pipelines split documents using fixed token windows or sentence-boundary heuristics \citep{lewis2020rag,gao2023rag}. TE sidesteps the chunking question: compression produces atomic fact lines as a structural by-product, so no separate chunking stage is needed. Conceptually adjacent are controlled natural languages such as Attempto Controlled English \citep{fuchs2008attempto}, but those are designed for theorem-proving rather than compression, and consumed by formal reasoners rather than LLMs.

\section{The Telegraph English Grammar}
\label{sec:grammar}

The grammar (version~5) lives in a 430-line specification document that doubles as the system prompt for the LLM-based compressor. We summarise its key design principles here; the full specification is supplementary material.

\subsection{Foundations}

Four principles govern the grammar, in strict priority order: (i)~\textbf{fidelity over brevity}---no information may be dropped unless inferable from what remains; (ii)~\textbf{atomic line structure}---each line contains exactly one claim, step, event, or question; (iii)~\textbf{upper-case default}, except where case carries information (proper names, code, SI symbols); (iv)~\textbf{target compression $\sim$5$\times$} when feasible, but correctness, auditability, and reversibility take strict priority over token reduction.

\subsection{Symbol vocabulary}

TE defines a fixed vocabulary of relational and logical operators. The full set numbers roughly 40; Table~\ref{tab:symbols} shows the core symbols that appear in most compressions. Each symbol has a single, non-interchangeable meaning. The grammar caps symbol density at three consecutive symbols per line---a readability constraint learned from early iterations where dense symbol chains became opaque even to GPT-4.

\begin{table}[h]
  \caption{Core relational and logical operators in the TE symbol vocabulary. The full vocabulary contains roughly 40 symbols organised by function (causal, logical, comparative, modal).}
  \label{tab:symbols}
  \centering
  \small
  \begin{tabular}{cll}
    \toprule
    Symbol & Meaning & Example \\
    \midrule
    \texttt{=} & Definition / equality & \texttt{VELOCITY=DISTANCE/TIME} \\
    \teArrow{} & Causation / flow & \texttt{HEAT\teArrow{}EXPANSION} \\
    \teImplies{} & Logical implication & \texttt{RAIN\teImplies{}WETNESS} \\
    \teTherefore{} & Therefore / conclusion & \texttt{X>Y \teAnd{} Y>Z \teTherefore{} X>Z} \\
    \teBecause{} & Because / reason & \texttt{MOTOR-FAILURE \teBecause{} OVERLOAD} \\
    \teUp{}/\teDown{} & Increase / decrease & \texttt{TEMPERATURE\teUp{}} \\
    \teAnd{}/\teOr{}/\teNot{} & And / or / not & \texttt{A\teAnd{}B},~~\texttt{\teNot{}EVIDENCE} \\
    \teApprox{}/\teNeq{} & Approximate / not equal & \texttt{COST\teApprox{}USD10M} \\
    \texttt{VS} & Contrast (never causal) & \texttt{MODEL-A VS MODEL-B} \\
    \bottomrule
  \end{tabular}
\end{table}

\subsection{Tags and domain conventions}

Beyond the symbol vocabulary, the grammar provides a tagging system and a set of domain-specific formatting rules. Tags handle the framing that natural language carries through verbose syntactic constructions: temporal state (\texttt{PAST:}, \texttt{NOW:}, \texttt{FUTURE:}), modality (\texttt{LIKELY:}, \texttt{POSSIBLE:}, \texttt{CONF=0.87}), roles (\texttt{AGENT:}, \texttt{PATIENT:}, \texttt{INSTRUMENT:}), scope (\texttt{CTX:} for shared context), and structured content types (\texttt{DEF:}, \texttt{Q:}/\texttt{A:}). Each tag does double duty: it collapses verbose framing into a single token and provides the structural handle that downstream systems use for selective retrieval and context management.

Domain conventions standardise the surface forms that vary most across writers: quantities (\texttt{VAR=VALUEUNIT}), citations (\texttt{[AUTH:YEAR]}, \texttt{DOI:}, \texttt{ARXIV:}), financial data (\texttt{USD10.5\,M}, \texttt{Y/Y+5\%}, \texttt{+2.5PT}), and URLs. Locking these down at the grammar level removes a class of factual-error failure modes the compressor would otherwise need to handle case by case.

\subsection{Compressor self-verification}

Two mechanisms keep the compressor honest within a single LLM call: a quality gate and a prescribed distillation sequence. The quality gate is a 12-point checklist covering formatting consistency, symbol precision, abbreviation policy, number formatting, information preservation, and citation integrity. It is embedded directly in the compression prompt, so the compressor self-verifies output before returning it. The distillation sequence prescribes a six-pass reasoning order: (1)~concept identification, (2)~claim extraction, (3)~relation mapping, (4)~redundancy elimination, (5)~numerical verification, (6)~citation cross-checking. This is a chain-of-thought scaffold inside a single inference, not a multi-call pipeline. Ordering matters: numerical verification before citation cross-checking, because citations sometimes attach to numerical claims that must be confirmed first.

\subsection{Compression as semantic chunking}

Compression and semantic chunking are not separate stages---they are the same operation. Every TE output line is an atomic fact, every section is tagged, and every \texttt{CTX:} block defines a scope. The structure falls out of the grammar's line-structure rule, not from any additional processing, and it enables three things that token-deleted text cannot support: \textbf{selective retrieval} (a query about adverse events retrieves exactly the relevant line and its scope, no chunking heuristic required); \textbf{graduated compression-on-read} (a context-assembly system can keep the most relevant lines at full fidelity, retain only heading tags for moderately relevant sections, and drop irrelevant sections entirely---no LLM call needed); and \textbf{continuous state refinement} (facts can be updated in-place, merged, or pruned during a session). We call this the \emph{compress-once, manage-continuously} principle. A worked example and a more detailed treatment appear in Appendix~\ref{app:semantic-chunking}.

\section{Experimental Setup}
\label{sec:setup}

\subsection{Dataset}

LongBench-v2 \citep{bai2024longbench} supplies the source corpus: 503 long-context documents. We filter to three categories suitable for factual QA---Single-Document QA, Multi-Document QA, and Long-Dialogue History Understanding---which leaves 339 documents. NLTK sentence tokenisation chunks each one into segments of at most 1{,}000 words, producing 4{,}081 chunk-level evaluation units. The categories span technical reports, multi-source narrative synthesis, and conversational history---three regimes where compression methods fail differently. The 1{,}000-word chunk cap matches the practical input size for which prompt compression actually saves money.

\subsection{Compression}

\begin{figure}[h]
  \centering
  \begin{minipage}[t]{0.48\linewidth}
    \centering
    \includegraphics[width=\linewidth]{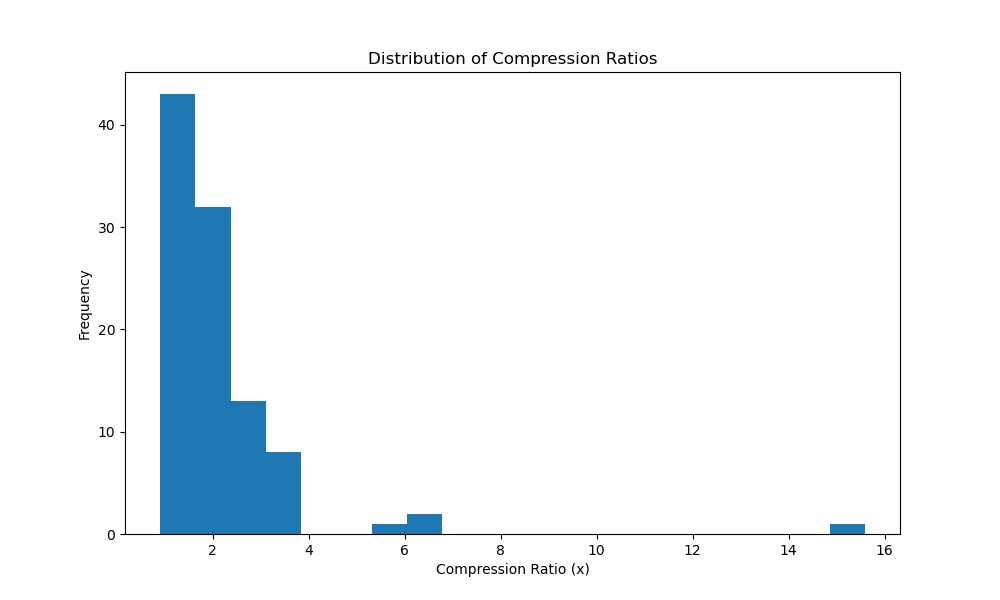}
    \caption{Distribution of compression ratios across 4{,}081 LongBench-v2 chunks compressed with TE (o4-mini, tiktoken cl100k\_base). Mean 0.585, range 0.13--1.57. The right tail above 1.0 corresponds to short, dense inputs that expand under TE.}
    \label{fig:ratio}
  \end{minipage}\hfill
  \begin{minipage}[t]{0.48\linewidth}
    \centering
    \includegraphics[width=\linewidth]{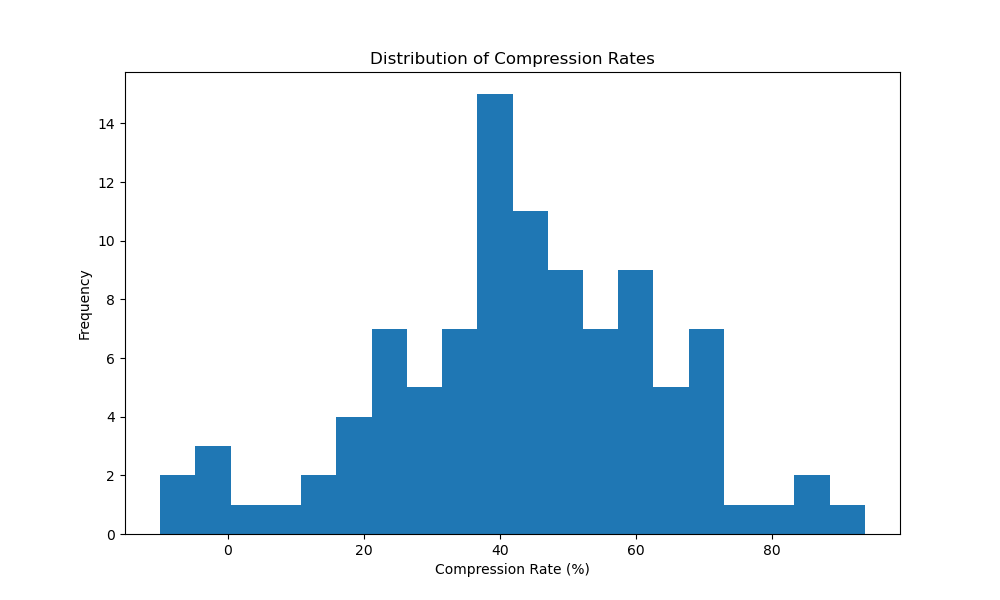}
    \caption{Distribution of per-chunk compression rate ($1-\textrm{ratio}$) over the same corpus. The median chunk loses roughly 43\% of its tokens; the bottom decile loses very little, reflecting TE's adaptive behaviour on already-dense inputs.}
    \label{fig:rate}
  \end{minipage}
\end{figure}

Each chunk is compressed into TE using the v5 grammar prompt with OpenAI's o4-mini model. Token counts are measured with tiktoken (cl100k\_base). The mean compression ratio is 0.585---a 41.5\% token reduction---with a range from 0.13 to 1.57. The upper end deserves explanation: rare, very short inputs that are already informationally dense occasionally \emph{expand} under TE, because the grammar's fidelity-first principle prohibits dropping information even when doing so would reduce token count. This is a feature, not a failure. The full distribution is shown in Figures~\ref{fig:ratio} and~\ref{fig:rate}.

For the LLMLingua-2 baseline, the same chunks are compressed using the publicly available \texttt{llmlingua} package at two retention rates: 0.50 (50\% kept) and 0.33 (33\% kept).

\subsection{QA evaluation protocol}

We design a multiple-choice protocol that isolates comprehension: can a model answer a factual question correctly when reading compressed text instead of the original? GPT-4.1 generates a verbatim QA pair from the original chunk, plus a semantically equivalent ``modified answer'' that prevents simple string matching from inflating scores. GPT-4.1 (temperature 0.7) generates three plausible distractors matched in style, length, and specificity. The modified answer and three distractors are shuffled into a four-option question. The evaluation model sees the original, then the compressed text, and selects an answer in each setting. Accuracy is the fraction of correct selections; an \emph{error} is a case where the model answered correctly on the original but incorrectly on the compressed version.

\subsection{Test suites and models}

Two suites probe different levels of information preservation. \textbf{key\_facts} (4{,}081 QA pairs) targets core concepts---headline findings, main claims, central arguments---with generically plausible distractors. \textbf{fine\_facts} (801 QA pairs) is adversarially designed to target information that lossy compression is most likely to destroy: precise numerical qualifiers, conditional statements, boundary conditions, secondary details. Distractors are near-miss variants---e.g.\ changing 4.8\% to 4.3\%---that can only be distinguished with access to the exact original detail.

We evaluate five OpenAI models: GPT-4.1, GPT-4o, GPT-4o-mini, GPT-4.1-nano, and a fine-tuned GPT-4o variant. GPT-4.1 also generates the QA pairs and distractors. Different suites use different model subsets: key\_facts is run on GPT-4.1, GPT-4o-mini, and GPT-4.1-nano; fine\_facts on GPT-4o and GPT-4o-mini. The fine-tuned variant is reported in the cost analysis (\S\ref{sec:cost}) but is not used as a separate accuracy benchmark---it serves as a sanity check that fine-tuning on the original distribution does not change comparative behaviour at compression-decoded inputs.

\section{Results}
\label{sec:results}

\subsection{Key facts accuracy}

\begin{table}[h]
  \caption{Accuracy on the \texttt{key\_facts} suite (4{,}081 QA pairs). TE is Telegraph English at $\sim$50\% compression; LLML2-50 is LLMLingua-2 at 50\% retention. Drop is in percentage points (pp) relative to original. \textbf{Bold} = best compressed.}
  \label{tab:key-facts}
  \centering
  \small
  \begin{tabular}{lccccc}
    \toprule
    Model & Original & TE & LLML2-50 & TE Drop & LLML2-50 Drop \\
    \midrule
    GPT-4.1       & 1.000 & \textbf{0.991} & 0.990 & $-0.9$ & $-1.0$ \\
    GPT-4o-mini   & 0.991 & \textbf{0.957} & 0.946 & $-3.4$ & $-4.5$ \\
    GPT-4.1-nano  & 0.980 & \textbf{0.950} & 0.949 & $-3.0$ & $-3.1$ \\
    \bottomrule
  \end{tabular}
\end{table}

On headline facts (Table~\ref{tab:key-facts}), TE matches or edges out LLMLingua-2 across the board. The accuracy loss is negligible for GPT-4.1---less than a percentage point while halving the token count. The gap widens on smaller models: 1.1 pp on GPT-4o-mini, with the same direction at GPT-4.1-nano. Not dramatic. But consistent---the direction never reverses across configurations.

\subsection{Fine facts accuracy}

\begin{table}[h]
  \caption{Accuracy on the adversarial \texttt{fine\_facts} suite (801 QA pairs). Fine-detail tasks expose larger compression effects; TE preserves more than LLMLingua-2 at matched retention.}
  \label{tab:fine-facts}
  \centering
  \small
  \begin{tabular}{lccccc}
    \toprule
    Model & Original & TE & LLML2-50 & TE Drop & LLML2-50 Drop \\
    \midrule
    GPT-4o      & 0.996 & \textbf{0.965} & 0.933 & $-3.1$ & $-6.3$ \\
    GPT-4o-mini & 0.938 & \textbf{0.843} & 0.820 & $-9.5$ & $-11.8$ \\
    \bottomrule
  \end{tabular}
\end{table}

Fine details are harder (Table~\ref{tab:fine-facts}). Compression loss runs $3$--$4\times$ higher than on key facts, regardless of method. TE holds an advantage of 3.2 pp over LLMLingua-2 on GPT-4o and 2.3 pp on GPT-4o-mini at matched 50\% retention. Against more aggressive LLMLingua-2 at 33\% retention (full numbers in Appendix~\ref{app:tables}), TE's lead grows to roughly 11 pp on GPT-4o-mini, where LLMLingua-2 drops a full 21 pp from baseline. That configuration is where token deletion starts to break down: it is removing the very tokens the questions probe.

\subsection{Accuracy hierarchy and compression statistics}

Across all models and tasks the ranking holds without exception: original $>$ TE $>$ LLML2-50 $>$ LLML2-33. TE's mean compression ratio of 0.585 ($\textrm{std}=0.254$) hides a wide spread: half the corpus sits between 0.41 and 0.74, with median 0.57. Documents dense with technical content or data tables resist compression; verbose narrative text yields ratios of 5:1 or better. Fidelity-first design means the ratio is an outcome, not a parameter.

\subsection{Error analysis}

Of the 4{,}081 key\_facts items, 187 (4.6\%) were correct on the original and incorrect on TE for GPT-4.1-nano. These error cases have a mean compression ratio of 0.531, slightly more compressed than the population mean---aggressive compression and error risk are correlated. Failures cluster around fine details: dates, units, conditional qualifications, and numerical relationships where TE either abbreviates a critical modifier or collapses a distinction the question specifically probes. One characteristic failure: a legal-document chunk where TE compressed ``no later than 30 calendar days after receipt of written notice'' into \texttt{DEADLINE=30D-AFTER-NOTICE}, and the question asked whether the deadline was in calendar or business days. The \texttt{30D} abbreviation does not distinguish. This is a limitation of the symbol vocabulary, not a compressor error.

\section{Analysis}
\label{sec:analysis}

\subsection{Why semantic rewriting outperforms token deletion}

Four mechanisms explain the pattern in the results. They are not ranked; different mechanisms dominate in different regimes. \textbf{Semantic-unit preservation}: token deletion operates at the token level and can split multi-word expressions, sever noun-modifier pairs, strand a number from its unit; TE works one level up, with related concepts grouped into hyphenated compounds and complete claims occupying single lines. \textbf{Explicit logical structure}: when LLMLingua-2 deletes a connective like ``therefore'' or ``in contrast to,'' the downstream model has to guess the relationship; TE refuses to offer the guess, with \teTherefore{}, \texttt{VS}, \teArrow{} each unambiguous and preserved regardless of what else is removed. \textbf{Co-reference stability}: TE's one-claim-per-line discipline and upper-case entity naming eliminate pronoun resolution ambiguity; token deletion can strand a pronoun whose antecedent has been removed. \textbf{Adaptive compression}: a fixed-ratio method compresses dense and verbose passages identically; TE does not---dense passages emerge at ratios near 1.0, verbose ones below 0.2. The four mechanisms interlock, which is why LLMLingua-2 cannot match TE by adopting any single one of them.

\subsection{The small-model effect}

The TE advantage grows as model capacity shrinks. GPT-4.1 barely notices the difference between TE and LLMLingua-2 on key facts; GPT-4.1-nano and GPT-4o-mini show a wider gap, and on fine facts the divergence becomes substantial. The likely explanation is capacity-dependent. Smaller models have less ability to reconstruct implicit relationships from token-deleted fragments; TE compensates by offloading that reconstruction work to the compression stage---the evaluation model receives a representation where the relationships are already marked, rather than having to hallucinate them from sparse clues. This has practical weight: smaller models are precisely the ones deployed in cost-sensitive production pipelines, which is where prompt compression earns its keep.

\subsection{The fine-facts gap}

Key facts survive both compression methods reasonably well. Central claims are often redundantly signalled, and even aggressive token deletion tends to preserve them. Fine details are stubborn in a different way: precise numerical qualifiers, conditional caveats, and secondary attributions are exactly the tokens an entropy-based classifier flags as low-importance in isolation. A number like ``4.8\%'' may look dispensable next to surrounding prose. But if the question asks whether the figure was 4.8\% or 4.3\%, that token is the entire answer. TE's claim-level decomposition and explicit numerical formatting (\texttt{+27.5\%}, \texttt{CONF=0.87}, \texttt{Y/Y+12.3\%}) are designed to preserve these details: numbers are never abbreviated, always attached to their units, and always placed in a structured format the downstream model can parse unambiguously.

\subsection{Pipeline-level cost}
\label{sec:cost}

There is a structural difference between the two methods that the accuracy comparison alone obscures: LLMLingua-2 operates as an input-only preprocessor. It compresses the initial prompt; generated output passes uncompressed to subsequent stages. TE can persist as a native format throughout a pipeline. Consider a five-step agent pipeline with 2{,}000 tokens of initial context and five generation steps averaging 400 tokens each, at \$10 per million tokens (Table~\ref{tab:cost}). The savings compound because each stage operates on TE-formatted text. LLMLingua-2 compresses only the first stage's input; the remaining four stages process uncompressed output at full token cost. A more architectural treatment of dynamic context management appears in Appendix~\ref{app:dynamic}.

\begin{table}[h]
  \caption{Pipeline-level cost for a five-step agent pipeline (2{,}000-token initial context, five 400-token generations) at \$10 per million tokens. TE persists across stages; LLMLingua-2 compresses only the first stage.}
  \label{tab:cost}
  \centering
  \small
  \begin{tabular}{lccc}
    \toprule
    Method & Total tokens & Cost / 1K calls & Savings \\
    \midrule
    Original          & 4{,}000 & \$40 & --- \\
    LLMLingua-2       & $\sim$3{,}300 & \$33 & \$7 \\
    Telegraph English & $\sim$1{,}600 & \$16 & \$24 \\
    \bottomrule
  \end{tabular}
\end{table}

\section{Implementation}

The reference implementation is a Python package with five pipeline stages: synchronous and asynchronous (Batch API) compression of LongBench-v2 documents using the TE grammar prompt; automated quality review via Claude (structured JSON scores 0--10 with strengths, weaknesses, and example pairs); end-to-end QA benchmarking (generation, distractor creation, MC evaluation); LLMLingua-2 baseline evaluation against the same QA pairs; and error analysis with case-level output. Each stage is accessible as both a CLI command and an importable library function.

\section{Limitations}

\textbf{LLM-dependent compression.} TE requires an LLM call per chunk, adding latency and cost at compression time. This is amortised when compressed text is reused, but TE is poorly suited for compressing ephemeral inputs that will be read once and discarded. \textbf{Proprietary evaluation models.} Our benchmark relies on OpenAI models that are not open-weight, limiting reproducibility; future work should extend evaluation to open models. \textbf{English only.} The grammar and benchmarks are English; adapting the symbol vocabulary to other languages---particularly agglutinative or logographic ones---is non-trivial. \textbf{Compressor model sensitivity.} TE quality depends on the model performing the rewrite; we have not yet mapped this sensitivity curve. \textbf{QA generation bias.} Both QA pairs and evaluations are produced by OpenAI models; an ideal evaluation would include human-written questions or a diverse set of QA generators. \textbf{Dynamic context management is not yet benchmarked.} The semantic chunking and dynamic state-management capabilities described in \S\ref{sec:grammar} and Appendix~\ref{app:dynamic} are architectural arguments, not empirical results from a multi-turn evaluation. We have demonstrated format-level feasibility; we have not measured downstream effects over extended sessions. This is the most important gap in the current evaluation. \textbf{Comparison scope.} We benchmark against LLMLingua-2 only---currently the strongest published baseline at our compression ratios. A broader comparison against AutoCompressors, RECOMP, and more recent methods would strengthen the claims.

\section{Conclusion}

Telegraph English demonstrates that structured semantic rewriting is a viable alternative to token deletion for prompt compression---and, on the evidence presented here, a better one. The advantage is largest where compression matters most practically: on smaller, cheaper models and on fine-grained details. The quantitative comparison may not be the most interesting part of this work. Token-deletion methods produce a smaller copy with no internal organisation; TE produces a representation where every line is an identified fact, every section is tagged, every relationship is marked with an explicit symbol. That structure makes the output not just smaller but more \emph{useful}---more retrievable, more auditable, more maintainable over time. The compress-once, manage-continuously principle is, at this stage, an architectural argument rather than an empirical result; validating it in production agent systems is the obvious next step.

\bibliographystyle{plainnat}
\bibliography{references}

\appendix

\section{Compression as Semantic Chunking: Worked Example}
\label{app:semantic-chunking}

To make \S\ref{sec:grammar} concrete, consider a multi-paragraph clinical-trial summary compressed into TE:

\begin{quote}\small\ttfamily
H1: CLINICAL-TRIAL OUTCOMES \\
CTX: PHASE-III RANDOMISED CONTROLLED-TRIAL(RCT); N=2400 \\
\hspace*{1em} PRIMARY-ENDPOINT: MORTALITY\teDown{}23\% VS PLACEBO; p<0.001 [SMITH:2024] \\
\hspace*{1em} SECONDARY-ENDPOINT: HOSPITALIZATION\teDown{}18\%; p=0.003 \\
\hspace*{1em} ADVERSE-EVENTS: NAUSEA=12\% \teAnd{} HEADACHE=8\% \teAnd{} SERIOUS=2.1\% \\
H1: SUBGROUP-ANALYSIS \\
\hspace*{1em} AGE>65: MORTALITY\teDown{}31\% (STRONGER-EFFECT) \\
\hspace*{1em} AGE<65: MORTALITY\teDown{}14\% (WEAKER-EFFECT) \\
\hspace*{1em} CONF=0.92 FOR INTERACTION-EFFECT \\
H1: LIMITATIONS \\
\hspace*{1em} FOLLOW-UP=18 MONTHS; LONG-TERM-EFFECTS UNKNOWN \\
\hspace*{1em} EXCLUSION: PATIENTS WITH RENAL-IMPAIRMENT
\end{quote}

Each line is a fact; each heading is a section boundary; each \texttt{CTX:} block defines a scope. The structure falls out of the grammar's line-structure rule rather than from any additional processing, and it enables three things that token-deleted text cannot support.

\paragraph{Selective retrieval.} A query about adverse events retrieves exactly the \texttt{ADVERSE-EVENTS} line and its \texttt{CTX:} scope. No sliding-window heuristic, no overlap parameter, no risk of splitting a relevant fact across chunk boundaries; the semantic boundaries are intrinsic to the format.

\paragraph{Graduated compression-on-read.} When assembling a prompt under a tight token budget, an agent can apply different policies to different sections: keep the lines most relevant to the current query at full fidelity; retain only the heading tags (\texttt{H1: LIMITATIONS}) for moderately relevant sections, preserving topic structure at near-zero cost; drop irrelevant sections entirely. This second-stage compression is semantically principled---it operates on identified sections, not on token positions.

\paragraph{Continuous state refinement.} During a conversation, facts from earlier turns can be revised without re-compressing the source: \emph{update} (replace a corrected figure in place), \emph{merge} (combine related facts when the distinction no longer matters), \emph{prune} (remove claims that have moved past relevance), and \emph{promote/demote} (expand a heading-collapsed section, or collapse a fully expanded one).

\section{Beyond Static Compression: Dynamic Context Architecture}
\label{app:dynamic}

The results in \S\ref{sec:results} measure TE as a static compression method---compress once, read once, evaluate. This is the fair comparison against LLMLingua-2 and where the benchmark numbers live. But the more consequential property of TE may not be the compression ratio; it is the structure of the output.

\paragraph{Unifying compression and chunking.}
Conventional RAG systems run documents through two stages: chunking (splitting into fixed-size segments for embedding) and optional compression (reducing each chunk's token count). These stages have different objectives and can interfere---a chunk boundary splits a sentence, then compression deletes the tokens needed to reconstruct it. TE collapses both stages into one. Each output line is a complete semantic unit; the chunking boundaries \emph{are} the compression output. A TE-compressed document is immediately embeddable at the line level. The practical consequence for retrieval precision: fixed-window chunking inevitably includes irrelevant context within each chunk and risks splitting relevant information; TE surfaces exactly the facts a query matches, at the granularity of individual claims.

\paragraph{Hierarchical context budgeting.}
Because TE output is tagged with headings, context scopes, and role markers, a context-assembly system can make graduated decisions about inclusion. For a given token budget: full-fidelity inclusion of all atomic lines for the most relevant sections; heading-only retention for moderately relevant sections, preserving topic structure at near-zero cost; omission of irrelevant sections entirely. This graduated policy can achieve very high total compression ($10$--$50\times$) when only a fraction of the document is relevant, while maintaining full detail where it matters. The policy operates on the TE output's structure---no LLM call needed.

\paragraph{Dynamic state in agentic sessions.}
Long-running agent sessions accumulate context over many exchanges. The standard solutions are blunt: hard truncation drops the oldest tokens regardless of relevance; periodic summarisation requires an LLM call and is irreversible. TE enables something finer. Because context is already decomposed into tagged atomic facts, an agent can maintain a living state: \emph{fact updates} replace the old line in place rather than appending alongside it; \emph{redundancy pruning} removes facts whose information has been absorbed by later ones; \emph{scope closure} collapses an entire \texttt{CTX:} block to a heading once a topic is resolved; \emph{priority re-ranking} reorders facts by current relevance, placing the most important context where transformer attention is strongest. Context growth is controlled by continuously refining the active fact set, not by discarding the oldest tokens. This is cheap (string manipulation, no LLM calls) and semantically principled.

The cost profile is asymmetric by design: one expensive LLM rewrite per document, then indefinite cheap manipulation of the structured output.

\section{Full Results Tables}
\label{app:tables}

\begin{table}[h]
  \caption{Complete \texttt{key\_facts} results with compression statistics.}
  \label{tab:a1}
  \centering
  \small
  \begin{tabular}{lcccccccc}
    \toprule
    Model & $n$ & Original & TE & LLML2-50 & TE Drop & LLML2-50 Drop & Mean ratio \\
    \midrule
    GPT-4.1       & 4{,}081 & 1.000 & 0.991 & 0.990 & $-0.9$ & $-1.0$ & 0.585 \\
    GPT-4o-mini   & 4{,}081 & 0.991 & 0.957 & 0.946 & $-3.4$ & $-4.5$ & 0.585 \\
    GPT-4.1-nano  & 4{,}081 & 0.980 & 0.950 & 0.949 & $-3.0$ & $-3.1$ & 0.585 \\
    \bottomrule
  \end{tabular}
\end{table}

\begin{table}[h]
  \caption{Complete \texttt{fine\_facts} results.}
  \label{tab:a2}
  \centering
  \small
  \begin{tabular}{lccccccc}
    \toprule
    Model & $n$ & Original & TE & LLML2-50 & TE Drop & LLML2-50 Drop \\
    \midrule
    GPT-4o      & 801 & 0.996 & 0.965 & 0.933 & $-3.1$ & $-6.3$ \\
    GPT-4o-mini & 801 & 0.938 & 0.843 & 0.820 & $-9.5$ & $-11.8$ \\
    \bottomrule
  \end{tabular}
\end{table}

\begin{table}[h]
  \caption{Compression ratio statistics (tiktoken cl100k\_base, $n=4{,}081$ chunks).}
  \label{tab:a3}
  \centering
  \small
  \begin{tabular}{lc}
    \toprule
    Statistic & Value \\
    \midrule
    Mean & 0.585 \\
    Std & 0.254 \\
    Min & 0.000 \\
    25th percentile & 0.407 \\
    Median & 0.570 \\
    75th percentile & 0.739 \\
    Max & 1.567 \\
    \bottomrule
  \end{tabular}
\end{table}

\begin{table}[h]
  \caption{Error analysis: \texttt{key\_facts} items correct on original, incorrect on TE (GPT-4.1-nano).}
  \label{tab:a4}
  \centering
  \small
  \begin{tabular}{lc}
    \toprule
    Statistic & Value \\
    \midrule
    Total items & 4{,}081 \\
    Error items & 187 \\
    Error rate & 4.58\% \\
    Mean compression ratio (errors) & 0.531 \\
    Mean compression ratio (all) & 0.585 \\
    \bottomrule
  \end{tabular}
\end{table}
\end{document}